\documentclass[12pt]{article}

\usepackage{amssymb}
\usepackage{amsmath}
\usepackage{amscd}
\usepackage{latexsym}
\usepackage{graphicx}

\usepackage{cite}

\newcommand{\be}{\begin{equation}}
\newcommand{\ee}{\end{equation}}

\newcommand{\dlt}{\delta}

\newcommand{\bt}{\beta}
\newcommand{\vp}{\varphi}

\newcommand{\al}{\alpha}
\newcommand{\ra}{\rightarrow}

\newcommand{\lbd}{\lambda}

\newcommand{\cH}{{\cal H}}

\newcommand{\rgl}{\rangle}
\newcommand{\lgl}{\langle}

\begin{document}

\begin{center}

{\Large{\bf Evolutionary Processes in Quantum Decision Theory} \\ [5mm]

Vyacheslav I. Yukalov$^{1,2}$ }  \\ [3mm]

{\it 
$^{1}$Bogolubov Laboratory of Theoretical Physics \\
Joint Institute for Nuclear Research, Dubna 141980, Russia \\ [3mm]

$^2$Department of Management, Technology and Economics \\
ETH Z\"urich, Swiss Federal Institute of Technology,
Z\"urich CH-8032, Switzerland } \\ [3mm]

{\bf E-mail}: yukalov@theor.jinr.ru 

\end{center}

\vskip 2cm

\begin{abstract}
The review presents the basics of quantum decision theory, with the emphasis on
temporary processes in decision making. The aim is to explain the principal points
of the theory. The difference of an operationally testable rational choice between 
alternatives from a choice decorated by irrational feelings is elucidated. 
Quantum-classical correspondence is emphasized. A model of quantum intelligence 
network is described. Dynamic inconsistencies are shown to be resolved in the frame
of the quantum decision theory.
\end{abstract}

\vskip 2cm

{\parindent=0pt
{\bf Keywords}: quantum decision theory; operationally testable choice; dual decision 
process; quantum-classical correspondence; quantum intelligence network  }  

\newpage

\section{Introduction}

In recent years, there has appeared high interest to the possibility of formulating 
decision theory in the language of quantum mechanics. Numerous references on this 
topic can be found in the books \cite{Khrennikov_1,Busemeyer_2,Haven_3,Bagarello_4} 
and review articles \cite{Yukalov_5,Agrawal_6,Sornette_7,Ashtiani_8}. This interest
is caused by the inability of classical decision theory \cite{Neumann_9} to comply 
with the behaviour of real decision makers, which requires to develop other approaches. 
Resorting to the techniques of quantum theory gives hopes for a better representation
of behavioral decision making. There are several variants of using quantum mechanics
for interpreting conscious effects. The aim of the present review is not the description
of the existing variants, which would need too much space and can be found in the cited
literature \cite{Khrennikov_1,Busemeyer_2,Haven_3,Bagarello_4,Yukalov_5,Agrawal_6,
Sornette_7,Ashtiani_8}, but a survey of the approach suggested by the author and his
coworkers. This approach was coined \cite{Yukalov_10} {\it Quantum Decision Theory} (QDT). 

In the present review, we limit ourselves by considering quantum decision theory, but we
do not touch other trends in the ramified applications of quantum techniques, such as
quantum approaches in physics, chemistry, biology, economics and finances, in quantum 
information processing, quantum computing, and quantum games. It looks evident that all 
those fields cannot be reasonably described in a single review.   

Although the theory of quantum games shares similarities with decision theory, however 
there exists an important difference between the standard treatment of quantum games 
\cite{Meyer_98,Eisert_99,Piotrowski_100,Landsburg_101,Guo_102} and the main idea of quantum 
decision theory presented in the review. In the theory of quantum games, one usually assumes 
that players are a kind of quantum devices following quantum rules \cite{Khan_103,Khan_104}. 
However in the approach of quantum decision theory \cite{Yukalov_10}, decision makers do not 
have to be quantum devices, moreover, they can be real human beings. The mathematics of QDT 
is analogous to the mathematics in the theory of quantum measurements, where an observer is 
a classical human being, while the observed processes are characterized by quantum laws. 
In QDT, quantum theory is a technical language allowing for the description of decision 
processes. Quantum techniques turn out to be a very convenient tool for characterizing 
realistic human decision processes incorporating rational-irrational duality, because 
quantum techniques are designed for taking into account the dual nature of quantum world, 
such as the particle-wave duality. The mathematical generalization of decision-making 
processes by including into them the rational-irrational duality is one of the main 
achievements of QDT. Summarizing, the specific features of QDT, distinguishing it from many 
variants of decision theory employing quantum techniques, are as follows.

(i) The mathematics of QDT is analogous to that used for characterizing quantum 
measurements, so that there is a direct correspondence between QDT and the theory
of quantum measurements. (ii) The approach is general not only in the description of
both decision theory and quantum measurements, but also in its mathematical formulation
allowing for the interpretation of various decision events or quantum measurements. 
(iii) The theory shows the difference between the decision making with respect to 
operationally testable events and the choice under uncertainty caused by irrational
subconscious feelings. (iv) Single decision makings as well as successive decision 
makings are described on equal footing. (v) Temporal evolution of decision processes 
is formulated. (vi) Quantum-classical correspondence is preserved explaining the 
relation between quantum and classical probabilities. (vii) No paradoxes of classical 
decision theory arise. (viii) Different dynamic decision inconsistencies find natural 
explanation in the frame of QDT. (ix) The theory can be applied to single decision makers 
and 
also to decision-maker societies forming quantum intelligence networks.   
      
Of course, not all above mentioned aspects will and can be considered in detail in 
the present review. To describe all of them would require a huge book. We shall 
concentrate on the principal points the theory is based on and will emphasize 
temporal effects that have not been sufficiently discussed in the previous publications. 
In order to make for the reader clear the basic ideas of the QDT, it is useful to 
present the foundations in several steps. First, we need to define the process of 
decision making in the case of operationally testable events. This introduces the way 
of calculating quantum probabilities in the case of projection valued measures. Then it 
is straightforward to generalize the consideration to dual processes, when a decision 
is made by taking into account rational as well as irrational sides, which involves the 
use of positive operator-valued measures. Next, the approach to describing a society of 
decision makers, forming a quantum intelligence network, is presented. Finally, several 
dynamic decision inconsistencies are considered as an illustration of how they can be 
naturally resolved in the frame of QDT.

\section{Operationally Testable Events}   

The mathematics of QDT is similar to that employed for treating quantum measurements.
Therefore each formula can be interpreted from two sides, from the point of view of 
quantum measurements or decision theory 
\cite{Yukalov_10,Yukalov_11,Yukalov_12,Yukalov_13}.
Below as a rule we follow the interpretation related to decision theory, only 
occasionally mentioning quantum measurements. Some parts of the scheme below can be 
familiar to quantum mechanics, but here we suggest the interpretation in the language 
of decision theory, which is needed for introducing the basic notions of QDT. These 
basic notions are introduced step by step in order to better demonstrate the logic 
of the QDT. We will not plunge into the details of numerous applications that can be 
found in the published literature. The main aim of this review is the explanation of 
the principal points of the approach, with the emphasis on the evolutionary side of 
decision processes.      
   
The typical illustration of decision making is based on the choice between given 
alternatives, say forming a set $\{A_n\}$, enumerated by the index $n=1,2,\ldots,N_A$.
The alternatives correspond to events that can be operationally tested. Each alternative 
is represented by a vector $|A_n\rgl$, where the bra-cket notation \cite{Neumann_14} is 
used. The vectors of alternatives pertain to a Hilbert {\it space of alternatives}
\be
\label{1}
\cH_A = {\rm span}\; \{\; | \; n \; \rgl \; \} \;   ,
\ee
being a closed linear envelope over an orthonormal basis. Decisions are taken by a 
subject whose mind is represented by a Hilbert space 
\be
\label{2}
\cH_S = {\rm span}\; \{\; | \; \al \; \rgl \; \}
\ee
that can be called the {\it subject space of mind}, or just the {\it subject space}. 
Thus the total space, where decisions are made, is the {\it decision space}
\be
\label{3}
 \cH = \cH_A \bigotimes \cH_S \;  .
\ee
This is a closed linear envelope over an orthonormal basis,
$$
 \cH = {\rm span}\; \{\; | \; n \al \; \rgl \equiv 
| \; n \; \rgl \otimes | \; \al \; \rgl \; \} \;  .
$$

The alternatives from the given set, represented by vectors $|A_n \rgl$, are assumed 
to be orthonormal to each other,
$$
\lgl \; A_m \; | \; A_n \; \rgl = \dlt_{mn} \;   ,
$$
which means that the alternatives are mutually exclusive. The alternative vectors 
are not required to form a basis. To each alternative, it is possible to put into 
correspondence the alternative operator
\be
\label{4}    
\hat P(A_n) =  | \; A_n \; \rgl \lgl \; A_n \; | \;  .
\ee
These operators enjoy the properties of projectors
$$
 \hat P(A_m) \hat P(A_n) = \dlt_{mn} \hat P(A_n) \; , 
\qquad
\sum_n \hat P(A_n) = 1 \; ,
$$
whose family forms a projection-valued measure. 

The state of a decision maker is represented by a statistical operator that is 
a semi-positive trace-one operator,
\be
\label{5}
{\rm Tr}_\cH \;\hat\rho(t) = 1 \;   ,
\ee
depending on time $t$ and acting on the space $\mathcal{H}$. In decision making, 
$\hat{\rho}$ is termed the decision-maker state or just the decision state. The pair 
$\{\mathcal{H}, \rho(t)\}$ is a {\it decision ensemble}. The probability of observing 
an event $A_n$ is 
\be
\label{6}
p(A_n,t) = {\rm Tr}_\cH \; \hat\rho(t) \hat P(A_n) \;  ,
\ee
which is uniquely defined for a Hilbert space of dimensionality larger than two 
\cite{Gleason_15}. Since $\hat{P}(A_n)$ acts on the space $\mathcal{H}_A$, 
probability (\ref{6}) can be rewritten as
\be
\label{7}
p(A_n,t) = {\rm Tr}_A \; \hat\rho_A(t) \hat P(A_n) \;   ,
\ee
where the trace is over the space $\mathcal{H}_A$, and 
\be
\label{8}
 \hat\rho_A(t) =  {\rm Tr}_S \; \hat\rho(t) = 
\sum_\al \lgl \; \al \; | \;\hat\rho(t)\; | \;\al \; \rgl \; ,
\ee
with the trace over the space $\mathcal{H}_S$. Note that probability (\ref{7}) can 
also be written as
$$
p(A_n,t) = {\rm Tr}_A \; \hat P(A_n)  \hat\rho_A(t) \hat P(A_n) \;   .
$$

Expression (\ref{6}), or (\ref{7}), is the predicted probability of observing an 
event $A_n$, or in decision theory, this is the predicted probability of choosing 
an alternative $A_n$. When comparing the predicted theoretical probabilities with 
empirical probabilities, the former are interpreted in the frequentist framework 
\cite{Bohr_16,Ballentine_17}. Thus $p(A_n,t)$ is the fraction of equivalent decision 
makers preferring the alternative $A_n$ at time $t$. This can be reformulated in 
the different way: $p(A_n,t)$ is the frequency of choosing the alternative $A_n$ 
by a decision maker, if the choice is repeated many times under the same conditions 
until the time $t$.  
 
We say that an alternative $A_i$ is stochastically preferred (or simply preferred) to
an alternative $A_j$ if and only if $p(A_i) > p(A_j)$, and the alternatives $A_i$ 
and $A_j$ are stochastically indifferent if and only if $p(A_i) = p(A_j)$.

\section{Single Decision Making}

Usually, one considers decision making as a process instantaneous in time, which 
is certainly a too rough modeling. Here we stress the importance of dealing with 
realistic decision making developing in time. The evolutionary picture saves us 
from different inconsistences arising in the toy model of instantaneous decisions 
\cite{Yukalov_11,Yukalov_18}. 

Suppose at the initial time $t = 0$ we plan to make a choice between the alternatives
from the given set. This implies a stage of preparation, when the decision maker is 
characterized by a state $\hat{\rho}(0)$. The evolution of the decision-maker state 
in time can be represented as
\be
\label{9}  
 \hat\rho(t) = \hat U(t) \;\hat\rho(0) \; \hat U^+(t) \;  ,
\ee
involving the evolution operator $\hat{U}(t)$. To keep the state normalization 
intact, the evolution operator has to be unitary,
$$
\hat U^+(t) \hat U(t) = 1 \;  .
$$
By employing a self-adjoint evolution generator $\hat{H}(t)$, the time transformation 
of the evolution operator can be written as the Lie differential equation 
\be
\label{10}
 \frac{d}{dt} \; \hat U(t) =  \hat H(t) \hat U(t) \; ,
\ee
with the initial condition
\be
\label{11}
 \hat U(0) = \hat 1 \; .
\ee

A very important clause is the requirement that the considered decision makers be 
well defined individuals, who do not drastically vary in time, otherwise there would 
be no meaning in predicting their decisions. In other words, the mental features of 
decision makers at one moment of time are to be similar to those at a different moment 
of time. Briefly speaking, we can say that the decision makers should have the property 
of self-similarity. This imposes a restriction on the form of the evolution generator. 
This restriction can be written as the commutator
\be
\label{12}
\left[ \; \hat H(t) , \; \int_0^t \hat H(t') dt' \right] = 0 \; .
\ee
Recall that in quantum theory the operator commuting with the evolution generator
is called the integral of motion. In our case, condition (\ref{12}) does not mean 
that the evolution generator does not depend on time, but it tells us that the 
properties of this operator are in some sense invariant with time, preserving the 
similarity of the decision makers at different moments of time. So, in decision theory, 
condition (\ref{12}) can be understood as the invariance of the properties of decision 
makers, that is, of the {\it decision-maker self-similarity}. In the theory of operator 
differential equations, commutator (\ref{12}) is termed the Lappo-Danilevsky condition 
\cite{Lappo_19}. Under this provision, the evolution operator, satisfying the evolution 
equation (\ref{10}), with the initial condition (\ref{11}), reads as
\be
\label{13}
 U(t) = \exp\left\{ - i \int_0^t \hat H(t') \; dt' \; \right\} \;  .
\ee
This form of the evolution operator satisfies the group properties that can be 
represented as the property of functional self-similarity \cite{Yukalov_90}.

Till now, we have not specified the choice of a basis for the decision space 
$\mathcal{H}$. It is convenient to accept as the basis that one composed of the 
eigenfunctions of the evolution generator, such that
\be
\label{14}
 \hat H(t)\; | \; n\al \; \rgl = E_{n\al}(t) \; | \; n\al \; \rgl \;  .
\ee
Generally, the eigenfunctions here can depend on time. However, there are two cases
when this dependence is either negligible or absent at all. One possibility is when 
the eigenfunctions vary with time much slower than the eigenvalues $E_{n\alpha}(t)$.
Then there exists a time horizon till which the time variation of the eigenfunction 
can be neglected, which is regulated by a kind of adiabatic conditions \cite{Yukalov_91}.
The other case is the already assumed Lappo-Danilevsky condition (\ref{12}) implying
the decision-maker self-similarity. It is easy to show that the Lappo-Danilevsky 
condition (\ref{12}) is equivalent to the validity of eigenproblem (\ref{14}) with
time-independent eigenfunctions. Exactly the same situation happens in quantum theory
in the case of nondestructive, or nondemolition, measurements \cite{Yukalov_92,Kampen_93,
Shao_94,Braginsky_95,Mozyrsky_96,Yukalov_97,Yukalov_61}.

This basis is complete, since the evolution generator is self-adjoint. Then
\be
\label{15}
\hat U(t)\; | \; n\al \; \rgl = U_{n\al}(t) \; | \; n\al \; \rgl \; ,
\ee
where
\be
\label{16}
 U_{n\al}(t) = \exp\left\{ - i \int_0^t E_{n\al}(t') \; dt' \; \right\} \;  .
\ee

In this way, we find the expression for the time-dependent predicted probability of 
choosing the alternative $A_n$ at time $t$, 
\be
\label{17}
 p(A_n,t) = \sum_{n_1 n_2} \;\sum_\al 
U_{n_1\al}(t) \lgl \; \al n_1 \; |\; \hat\rho(0) \; | \; n_2\al \; \rgl \;
U^*_{n_2\al}(t) \lgl \;  n_2 \; |\; \hat P(A_n) \; | \; n_1 \; \rgl \; .
\ee
Again, to compare this expression with experimentally observed, it is reasonable 
to interpret it as a frequentist probability \cite{Bohr_16,Ballentine_17}, although 
other interpretations are admissible \cite{Fay_20}. 

The evolution generator is defined on the decision space $\cH$ and characterizes
the velocity of the change of the decision-maker state caused by the influence of 
the set of alternatives. The impact of the evolution generator on the decision-maker 
state is assumed to be finite, such that, if the decision process starts at time 
$t_A$ and ends at time $t_A+\tau_A$, being taken in the time interval 
$[t_A, t_A+\tau_A]$, then 
\be
\label{18}
 \left| \; \int_{t_A}^{t_A+\tau_A} E_{n\al}(t) \; dt \; \right|  < \infty \; .
\ee
Although this impact can be asymptotically large, remaining finite. The speed of 
the decision process can be quantified by a rate parameter $g$, which the eigenvalue 
$E_{n\alpha}(t) = E_{n\alpha}(t,g)$ depends on in such a way that in the limit of a
slow process 
\be 
\label{19}
E_{n\al}(t,g) \ra 0 \qquad ( g \ra 0 ) \;   ,
\ee
while under a fast process
\be
\label{20}
 E_{n\al}(t,g) \ra \infty \qquad ( g \ra \infty ) \;   .
\ee

In the case of a slow process, 
$$
U_{n\al}(t) \simeq 1 \qquad ( g \ra 0 ) \;  .
$$
This yields the probability
\be
\label{21}
 p(A_n,t) \simeq {\rm Tr}_A \; \hat\rho_A(0) \hat P(A_n) \qquad ( g \ra 0 ) \;  ,
\ee
which implies that the decision state practically does not change:
\be
\label{22}    
 \hat\rho_A(t)  \simeq  \hat\rho_A(0) \qquad ( g \ra 0 ) \; .
\ee

When the process is fast, in the sum (\ref{17}) one has
$$
U_{m\al}(t)\; U_{n\al}^*(t)  \simeq \dlt_{mn} \qquad ( g \ra \infty) \; .
$$
This leads to the probability
\be
\label{23}
 p(A_n,t) \simeq {\rm Tr}_A \; \hat\rho_A(t) \hat P(A_n) \qquad
 ( g \ra \infty) \;   ,
\ee
with the state
\be
\label{24}
 \hat\rho_A(t) \simeq \sum_m \hat P_m \; \hat \rho_A(0) \; \hat P_m 
\qquad
  ( g \ra \infty) \;  ,
\ee
where
$$
\hat P_m \equiv | \; m \; \rgl \lgl \; m \; |  \;  .
$$

In the intermediate cases of the process rate, one has to resort to the general
probability (\ref{17}).

\section{Changing Initial Conditions}

Assume that a subject made a decision during the time interval 
$[t_A, t_n\equiv t_A+\tau_A]$ choosing the alternative $A_n$. Then what would be 
the following evolution of the probability $p(A_n,t)$? Strictly speaking, there is 
nothing special in the fact of one subject making a decision at any moment of time. 
According to the frequentist understanding, the probability gives a distribution 
over an ensemble of decision makers or over many repeated decisions. This means that 
the probability yields a fraction of decision makers (or the fraction of decisions) 
preferring this or that alternative. In the following moments of time, the probability 
will continue to be defined by expression (\ref{17}). 

However, one can put the question in a different manner: Suppose we are interested 
in the decision making of just a single subject who certainly chose the alternative 
$A_n$, so that at the moment $t_n$ the probability became one, instead of that 
prescribed by expression (\ref{17}). Then what would be the following evolution of 
the probability? Again, there is nothing extraordinary in that case. Before making 
a decision, the predicted probability was described by expression (\ref{17}). After 
making the decision, if we insist that a posteriori probability became one, this means 
that we have to replace $p(A_n,t_n)$ by one, treating the latter as a new initial 
condition. So, the replacement
\be
\label{25}
p(A_n,t_n)  \longmapsto 1
\ee
means nothing but the change of the initial condition for the equation describing the
evolution of the decision state. The related replacement 
\be
\label{26}
\hat\rho_A(t_n) \; \longmapsto  \; \hat\rho_L(t_n)
\ee
assumes that starting from the moment of time $t_n$, we treat as the new initial 
condition the state defined by the equality
\be
\label{27}
{\rm Tr}_A \; \hat\rho_L(t_n) \hat P(A_n) = 1 \;   .
\ee
The simplest solution for the latter equation is the L\"uders state \cite{Luders_21}
\be
\label{28}
 \hat\rho_L(t_n) = \frac{\hat P(A_n) \hat\rho_A(t_n) \hat P(A_n)}
{{\rm Tr}_A \; \hat\rho_A(t_n) \hat P(A_n)} \; .
\ee
Note that here $\hat{\rho}_L(t_n)$ is the a posteriori post-decision state, while
$\hat{\rho}_A(t_n)$ is a priori anticipated state, that is
$$
 \hat\rho_L(t_n) \equiv \hat\rho_L(t_n+0) \; , \qquad
 \hat\rho_A(t_n) \equiv \hat\rho_A(t_n-0) \;   .
$$

The replacement (\ref{26}) often is labeled "collapse". For wave functions, this 
replacement is equivalent to the sudden change from a state $|\psi\rgl$ to the 
state $|A_n\rgl$. It is really easy to check that the wave function "collapse" 
implies the replacement
$$
\hat\rho_A(t_n) = | \; \psi \; \rgl \lgl \; \psi \; |  \;\; \longmapsto \;\;
\hat\rho_L(t_n) = | \; A_n \; \rgl \lgl \; A_n \; |  \; .
$$
Under the name "collapse" one means a sudden jump of the state, which could be 
dramatic provided the wave function or decision state would describe real matter. 
However, a decision state, as well as a wave function, are nothing but the 
probabilistic characteristics that can be used for calculating and predict a priori 
quantum probabilities satisfying evolution equations. Fixing a decision state at some 
moment of time simply means fixing new initial conditions for the state evolution. 
Thus the L\"uders state is just the new initial condition for the moment of time $t_n$, 
\be
\label{29}
 \hat\rho_L(t_n) = {\rm Tr}_S \; \hat\rho(t_n,t_n) \; ,
\ee
after which again the predicted probabilities have to be found from the state evolution
\be
\label{30}
\hat\rho(t,t_n) = \hat U(t,t_n) \;  \hat\rho(t_n,t_n) \; \hat U^+(t,t_n) \; ,
\ee
with the evolution operator
\be
\label{31}
\hat U(t,t_n) = \exp\left\{\; -i \int_{t_n}^t \hat H(t') \; dt' \; \right\} \; .
\ee
This defines the decision state
\be
\label{32}
\hat\rho_A(t,t_n) = {\rm Tr}_S \; \hat\rho(t,t_n) \qquad ( t > t_n)
\ee
after the time $t_n$. Respectively, this decision state gives the probability of 
choosing an alternative $A_m$ after the time $t_n$,
\be
\label{33}
p(A_m,t) = {\rm Tr}_A \; \hat\rho_A(t,t_n) \; \hat P(A_m) \qquad ( t > t_n) \;  .
\ee
Comparing expressions (\ref{29}) and (\ref{32}) yields the initial condition
\be
\label{34}
 \hat\rho_A(t_n,t_n) = \hat\rho_L(t_n) \; .
\ee

Following the same steps as in the previous section, we get the time evolution of the 
probability for $t > t_n$,
$$
p(A_m,t) = \sum_{n_1 n_2} \; \sum_\al 
U_{n_1\al}(t,t_n) \lgl \; \al n_1 \; | \; \hat\rho(t_n,t_n) \; | \; n_2 \al \; \rgl 
\times
$$
\be
\label{35}
\times
U^*_{n_2\al}(t,t_n) \lgl \;  n_2 \; | \; \hat P(A_m) \; | \; n_1 \; \rgl 
\qquad
(t > t_n ) \; ,   
\ee
where
$$
U_{n\al}(t,t_n) = \exp\left\{ \; - i \int_{t_n}^t E_{n\al}(t') \; dt' \; \right\} \;  .
$$
In the limit of a slow process, this results in
\be
\label{36}
  \hat\rho_A(t,t_n) \simeq \hat\rho_L(t_n) \qquad ( g \ra 0 ) \; .
\ee
And under a fast process, we have
\be
\label{37}
 \hat\rho_A(t,t_n) \simeq \sum_m \hat P_m \; \hat\rho_L(t_n) \; \hat P_m 
\qquad 
( g \ra \infty ) \;  .
\ee

Let us consider the situation, when a subject, after fixing the initial condition 
(\ref{29}), where the alternative $A_n$ was chosen, is interested in finding the 
probability of deciding on the alternative $A_m$. In the slow-process limit, we get
\be
\label{38}
 p(A_m,t) = {\rm Tr}_A \; \hat\rho_L(t_n)\; \hat P(A_m) \equiv  p_L(A_m,t_n) \;  .
\ee
Introducing the Wigner \cite{Wigner_22} probability
\be
\label{39}
p_W(A_m,t_n) = {\rm Tr}_A \; \hat P(A_n) \; \hat\rho_A(t_n)\; \hat P(A_n) \; \hat P(A_m)
\ee
yields the relation
\be
\label{40}
p_W(A_m,t_n) = p_L(A_m,t_n) \; p(A_n,t_n) \; ,
\ee
where
$$
p(A_n,t_n) = {\rm Tr}_A \; \hat\rho_A(t_n)\; \hat P(A_n)  \; .
$$

This equation reminds us the relation between the classical joint and conditional
probabilities. Therefore sometimes one is tempted to interpret the Wigner probability, 
$p_W(A_m,t_n)$ as a joint probability of two events, $A_n$ and $A_m$, while the 
L\"uders probability $p_L(A_m,t_n)$, as a conditional probability of the event $A_m$
under the event $A_n$ previously happened. This interpretation, however, cannot pretend
to provide a generalization of classical probabilities to quantum theory. This, first 
of all, because the derived relation is a very particular case of extremely slow processes,
when $g \ra 0$, but not the general expression. Second, this relation is not valid for all 
times, but it only connects the terms at time $t_n$. What is more important, this relation 
contains the terms taken from different sides of $t_n$, that is, connecting the predicted 
a priori probability with the post-decision a posteriori probability,
$$
p_L(A_m,t_n+0) = \frac{p_W(A_m,t_n-0)}{p(A_n,t_n-0)} \;  .
$$
But the meaningful definition of a probability should be valid for any time $t > t_n$.   
Third, the generalization of a classical expression assumes to contain the classical one 
as a particular case. But it is easy to check that the L\"{u}ders probability is
$$
p_L(A_m,t_n) = | \; \lgl \; A_m \; | \; A_n \; \rgl \; |^2 \; ,
$$   
which is symmetric with respect to the interchange between $A_m$ and $A_n$. For 
commuting events, corresponding to the classical case, the L\"{u}ders probability 
becomes trivial $\delta_{mn}$. Contrary to this, the classical conditional probability 
is neither symmetric nor trivial, which is confirmed by numerous empirical observations 
\cite{Boyer_23,Boyer_24}. In the best case, the L\"{u}ders probability is just a transition 
probability \cite{Yukalov_11,Yukalov_18}. 

Sometimes one tries to save the interpretation of the L\"uders probability as a conditional 
probability by resorting to the assumption of degenerate quantum states. This, nevertheless, 
does not save the situation because of several reasons: The origin of degeneracy is never 
defined, which makes the assumption groundless. The degeneracy is not important, since it 
always can be lifted by infinitesimally small variations of the problem 
\cite{Yukalov_11,Yukalov_18}. Finally, the degeneracy does not make classical expressions 
a particular case of quantum formulae, i.e. there is no quantum-classical correspondence 
\cite{Yukalov_11,Yukalov_18,Yukalov_25}.     
      
As we see, the consideration of realistic, developing in time, decision processes helps
us to avoid misrepresentation of the obtained expressions. The use of formal relations,
neglecting evolutionary processes, can lead to meaningless results like negative and 
complex probabilities \cite{Yukalov_11,Yukalov_18,Johansen_26}.

\section{Successive Decision Making}

When we consider the alternatives from the same set, say $\{A_n\}$, the probability of 
any $A_n$ is described by the approach elucidated in the previous sections, which 
leads to the set of the probabilities $\{p(A_n,t): n = 1,2,\ldots,N_A\}$. A rather 
different problem is the definition of successive decisions with respect to alternatives 
from different sets. Below we consider this problem employing some of the methods from
the theory of quantum measurements \cite{Neumann_14,Johansen_26,Johansen_27,Kalev_28},
however essentially modifying them in order to adjust to quantum decision theory 
\cite{Yukalov_11,Yukalov_18}.  

Suppose there are two sets of alternatives, $\{A_n: n = 1,2,\ldots,N_A\}$ and 
$\{B_k: k = 1,2,\ldots,N_B\}$. At the time interval $[t_A, t_A + \tau_A]$, a subject 
makes a decision with respect to the alternatives $A_n$. Then, in the time interval 
$[t_B + \tau_B]$, where $t_B > t_A + \tau_A$, the subject decides about $B_k$. 

According to the general approach of defining quantum joint probabilities 
\cite{Yukalov_11,Yukalov_18}, each set of alternatives is represented by the related 
set of alternative vectors and alternative projection operators,
$$
A_n \;\; \longmapsto \;\; |\; A_n \; \rgl \;\; \longmapsto \;\; 
\hat P(A_n) \equiv |\; A_n \; \rgl \lgl \; A_n \; | \; ,
$$
\be
\label{41}
B_k \;\; \longmapsto \;\; |\; B_k \; \rgl \;\; \longmapsto \;\; 
\hat P(B_k) \equiv |\; B_k \; \rgl \lgl \; B_k \; | \; .
\ee
Respectively, there are two spaces of alternatives:
\be
\label{42}
\cH_A= {\rm span}\{ |\; n \; \rgl \} \; , \qquad 
\cH_B= {\rm span}\{ |\; k \; \rgl \} \; ,
\ee
whose bases, in general, are different. Recollecting the existence of the subject space 
of mind $\mathcal{H}_S$, we have the decision space
\be
\label{43}
\cH = \cH_A \bigotimes  \cH_B \bigotimes  \cH_S  \; .
\ee

The decision-maker state is subject to the time evolution (\ref{9}). The evolution 
operator satisfies the evolution equation (\ref{10}), now with the evolution generator
\be
\label{44}
 \hat H(t) =  \hat H_{AS}(t) \bigotimes \hat 1_B \; + \; 
\hat 1_A \bigotimes \hat H_{BS}(t) .
\ee
Again it is convenient to choose the basis composed of the eigenfunctions of the 
evolution generator, such that
\be
\label{45}
  \hat H(t) \; |\; n k \al \; \rgl  = E_{nk\al}(t) \; | \; nk\al \; \rgl \; .
\ee
This assumes the self-similarity of the decision maker in the form of the Lappo-Danilevsky 
condition (\ref{12}), because of which we have
\be
\label{46}
 \hat U(t) \; |\; n k \al \; \rgl  = U_{nk\al}(t) \; | \; nk\al \; \rgl \; ,
\ee
where
\be
\label{47}
 U_{nk\al}(t) = \exp\left\{\; - i \int_0^t E_{nk\al}(t')\; dt' \; \right\} \; .
\ee

The joint probability that, first, a decision on $A_n$ is taken (an event $A_n$ happens)
and later a decision on $B_k$ is made (an event $B_k$ occurs) is defined as
\be
\label{48}
p(B_k A_n,t) \equiv 
{\rm Tr}_\cH \; \hat\rho(t) \; \hat P(B_k) \bigotimes \hat P(A_n) \; .
\ee
This also can be written as
\be
\label{49}
p(B_k A_n,t) \equiv 
{\rm Tr}_{AB} \; \hat\rho_{AB}(t) \; \hat P(B_k) \bigotimes \hat P(A_n) \; ,
\ee
where
\be
\label{50}
\hat\rho_{AB}(t) \equiv {\rm Tr}_S \; \hat\rho(t) = 
\sum_\al \lgl \; \al\; | \; \hat\rho(t) \; | \; \al\; \rgl \;  .
\ee
Expanding this results in the expression for the sought joint probability of choosing 
first the alternative $A_n$ and later the alternative $B_k$ yields
$$
p(B_k A_n,t) = \sum_{n_1 n_2} \;  \sum_{k_1 k_2} \; \sum_\al \; U_{n_1 k_1\al}(t) 
\lgl \; \al k_1 n_1 \; | \; \hat\rho(0) \; | \; n_2 k_2 \al \; \rgl \;
U^*_{n_2 k_2\al}(t) \times
$$
\be
\label{51}
\times \lgl \; k_2 \; | \; \hat P(B_k) \; | \; k_1 \; \rgl
\lgl \; n_2 \; | \; \hat P(A_n) \; | \; n_1 \; \rgl \; .
\ee
  
The process of taking decisions is supposed to be smooth, such that its impact does not
result in finite-time divergences,
\be
\label{52}
 \left| \; \int_{t_A}^{t_A+\tau_A} E_{nk\al}(t) \; dt \; + \;
\int_{t_B}^{t_B+\tau_A} E_{nk\al}(t) \; dt \; \right| \; < \; \infty \; .
\ee
The quantity $E_{nk\alpha}(t)$ characterizes the speed of the process of the subject 
deliberating about the given alternatives. For convenience, it is possible to define 
a rate parameter, entering the eigenvalue
\be
\label{53}
E_{nk\al}(t) = E_{nk\al}(t,g)
\ee
in such a way that a slow process would imply
\be
\label{54}
E_{nk\al}(t,g) \ra 0 \qquad ( g \ra 0 ) \;   ,
\ee
while a fast process would mean that
\be
\label{55}
E_{nk\al}(t,g) \ra \infty \qquad ( g \ra \infty) \;   .
\ee

When the process is slow, we have
\be
\label{56}
 U_{nk\al}(t) \simeq 1 \qquad ( g \ra 0 ) \;   .
\ee
Then the corresponding probability becomes
\be
\label{57}
 p(B_k A_n,t)  \simeq 
{\rm Tr}_{AB} \; \hat\rho_{AB}(0) \; \hat P(B_k) \bigotimes \hat P(A_n) 
\qquad
( g \ra 0 ) \; ,
\ee
while the decision state reads as
\be
\label{58}
 \hat\rho_{AB}(t) \simeq \hat\rho_{AB}(0) \qquad ( g \ra 0 ) \; .
\ee

It is useful to stress that even if at the initial moment of time the decisions on $A_n$
and $B_k$ formally look to be not explicitly connected, so that
$$
\hat\rho(0) =  \hat\rho_{AS}(0) \bigotimes \hat\rho_{BS}(0) \;  ,
$$
anyway the decision state does not factorize, 
$$
\hat\rho_{AB}(0)  = {\rm Tr}_S\; \hat\rho_{AS}(0) \bigotimes \hat\rho_{BS}(0) \; ,
$$
being entangled through the subject mind. Only in the extreme, little realistic, case 
when at the initial moment of time all processes are absolutely not correlated, so that
$$
\hat\rho(0) =  
\hat\rho_{A}(0) \bigotimes \hat\rho_{B}(0) \bigotimes \hat\rho_{S}(0) \; ,
$$
only then the joint probability factorizes,
$$
 p(B_k A_n,0) =  p(B_k,0) \; p(A_n,0) \; .
$$
Thus, even deciding on different alternatives, the total decision state, generally, 
remains entangled, therefore producing entangled states in the process of decision making
\cite{Yukalov_29}. The produced entanglement can be measured 
\cite{Yukalov_30,Yukalov_31,Yukalov_32,Yukalov_33,Yukalov_34,Yukalov_35}.      

For a fast process, we have
\be
\label{59}
 U_{n_1 k_1\al}(t)\; U_{n_2 k_2\al}(t) \simeq \dlt_{n_1 n_2} \dlt_{k_1 k_2} \;  .
\ee
Then the decision state reduces to
\be
\label{60}
\hat\rho_{AB}(t)  \simeq \sum_{n k} 
\hat P_n \bigotimes \hat P_k \; \hat\rho_{AB}(0) \; \hat P_n \bigotimes \hat P_k 
\qquad 
( g \ra \infty) \; ,
\ee
where
$$
\hat P_n = | \; n \; \rgl \lgl \; n \; | \; , \qquad 
\hat P_k = | \; k \; \rgl \lgl \; k \; | \; \; .
$$

It is important to emphasize that the joint probability (\ref{48}) is real-valued,
\be
\label{61}
p^*(B_k A_n,t) =  p(B_k A_n,t) \; .
\ee
It is not symmetric with respect to the interchange of $A_n$ and $B_k$, since the 
evolution generator (\ref{44}) depends on the order of decisions, first being $A_n$ and 
second, $B_k$. As is postulated at the beginning of this section, at the time interval 
$[t_A, t_A + \tau_A]$, a subject makes a decision on the alternatives $A_n$. Then, 
in the time interval $[t_B + \tau_B]$, where $t_B > t_A + \tau_A$, the subject decides 
on $B_k$. This means that the evolution generator can be presented as
\begin{eqnarray}
\nonumber
\hat H(t) = \left\{ \begin{array}{ll}
\hat H_{AS}(t) \bigotimes \hat 1_B \; , ~ & ~ t \leq t_A + \tau_A \\
\\
 \hat 1_A \bigotimes \hat H_{BS}(t) \; , ~ & ~ t > t_B > t_A + \tau_A
\end{array} \right. \; .
\end{eqnarray}
From here its dependence on the order of decisions is evident. We may notice that the
interchange of the decisions on $A_n$ and $B_k$ is similar to the inversion of time, 
hence
$$
 p(B_k A_n,t) =  p(A_n B_k,-t) \;  .
$$

Also, it is useful to compare the joint probability (\ref{49}) with the Kirkwood
distribution \cite{Kirkwood_36}. The latter is defined for two coinciding in time events,
whose projectors pertain to the same space of alternatives $\mathcal{H}_A = \mathcal{H}_B$,
which gives
$$
p_K(B_k A_n) \equiv {\rm Tr}_A \;\hat\rho_A \; \hat P(B_k) \;\hat P(A_n) \;  .
$$
It easy to see from the complex conjugate expression
$$
p_K^*(B_k A_n) = {\rm Tr}_A \;\hat\rho_A \; \hat P(A_n) \;\hat P(B_k) = p_K(A_nB_k)
$$
that 
$$
p_K^*(B_k A_n) \neq p_K(B_k A_n) \;   ,
$$
which tells us that the Kirkwood distribution, generally, is complex-valued. Thus the 
complex Kirkwood distribution and the real joint probability (\ref{49}) are principally 
different.

\section{Dual Decision Process}

Subjects rarely make decisions being based solely on the usefulness of alternatives, as 
prescribed by utility theory \cite{Neumann_9}, especially when decisions are to be made
under uncertainty. The choice between alternatives is practically always not purely
objective and based on well defined values that could be quantified, but strong subjective
feelings, biases, and heuristics are involved in the process of taking decisions
\cite{Tversky_37,Kahneman_38}. Intuition and emotions are also subjective, but they help 
us to take decisions, being a kind of decision-making process \cite{Minsky_39}. Taking 
account of subjective feelings and emotions in taking decisions is important for the 
problem of human-computer interaction \cite{Picard_40} and creation of artificial 
intelligence \cite{Yukalov_41,Yukalov_42}.

Understanding that human decision making involves two sides, the rational evaluation of 
utility and intuitive irrational attraction or repulsion towards each of the alternatives,
has been thoroughly investigated and elucidated in the approach called dual-process theory
\cite{Sun_43,Paivio_44,Stanovich_45,Kahneman_46}. Quantum techniques seem to be the most
appropriate for portraying the duality of human decision processes, since quantum theory 
presupposes the so-called wave-particle duality. Below we show how the dual-process 
approach \cite{Sun_43,Paivio_44,Stanovich_45,Kahneman_46} is formulated in quantum 
language \cite{Yukalov_10,Yukalov_30,Yukalov_47,Yukalov_48,Yukalov_49}. 

Thus to describe decision making by real subjects it is necessary to take into account 
the {\it rational-irrational} duality of decision making. First, we need to say several 
words on the meaning of the notion "rational" that can be understood in different senses 
\cite{Black_50}. It is necessary to distinguish between the philosophical and psychological 
meanings of this term. In philosophical writings, one gives the definition of "rational" 
as all what leads to the desired goal \cite{Searle_51}. However, under that definition, 
any illogical uncontrolled feeling that leads to the goal should be named rational. In 
decision making one uses the {\it psychological} meaning of the term "rational" as what 
can be explained and evaluated logically and what follows explicitly formulated rules. 
On the contrary, emotions, intuition, and moral feelings cannot be logically and 
explicitly formulated and quantified, especially at the moment of taking a decision. So, 
"rational" in decision making is what can be explicitly formulated, based on clear rules, 
deterministic, logical, prescriptive, normative. While "irrational" is just the opposite 
to "rational".

In this way, one has to separate the psychological and philosophical definitions of 
"rational" or "irrational". The psychological definition assumes that the distinction 
between "rational" and "irrational" is done at the moment of taking a decision. Conversely, 
the philosophical definition of "rational" as what leads to the goal is a definition that 
can work better afterwards, when the goal has been reached. Only then it becomes clear 
what was leading to the goal and what was not. At the moment of taking a decision, it is not 
always clear what leads to the goal and what does not.

The separation in decision making into rational and irrational has a simple and well 
defined psychological meaning based on actual physiological processes in the brain 
\cite{Ariely_52}. One often calls rational processes as conscious, while irrational 
processes as subconscious. 

From the mathematical point of view, the said above can be formulated as follows. Let us 
consider the choice between a set of alternatives $A_n$. An alternative is represented by 
a vector $|A_n\rgl$ in a Hilbert space of alternatives $\mathcal{H}_A$ and by a projector 
$\hat{P}(A_n)$ acting on this space. The subject space of mind is $\mathcal{H}_S$. So, the 
decision space is
\be
\label{62}
\cH = \cH_A \bigotimes \cH_S \; .
\ee

As is explained above, when choosing between alternatives, in addition to the rational
evaluation of the utility of each alternative, the decision maker experiences irrational
feelings. This implies that each alternative $A_n$, and its representation $|A_n\rgl$ in 
the space of alternatives $\mathcal{H}_A$, is complimented by a set of subjective feelings 
$z_n$ represented by a vector $|z_n \rgl$ in the subject space of mind $\mathcal{H}_S$. 
That is, in the process of decision making one compares not merely the alternatives $A_n$,
but the composite prospects
\be
\label{63}
\pi_n \equiv A_n z_n \; .
\ee

Being a member of the subject space of mind, the vector $|z_n \rgl$ allows for the 
expansion
\be
\label{64}
| \; z_n \; \rgl = \sum_\al b_{n\al} \; | \; \al \; \rgl \;  ,
\ee
in which $b_{n\alpha}$ are random quantities, which signifies a non-deterministic character 
of irrational feelings and emotions. 

We do not impose excessive number of conditions, trying to limit ourselves by the minimal 
number of restrictions. The required conditions will be imposed on the resulting expressions 
describing probability. Therefore the vectors $|z_n \rgl$ are not forced to be obligatory 
normalized, so that the scalar product
$$
 \lgl \; z_n \; | \; z_n \; \rgl = \sum_\al \; | b_{n\al}\; |^2
$$
does not have to be one. Also, the vectors $|z_n \rgl$ with different labels are not 
orthogonal to each other. Respectively, the operator
\be
\label{65}
\hat P(z_n) \equiv  | \; z_n \; \rgl \lgl \; z_n \; | = 
\sum_{\al\bt} b_{n\al}b^*_{n\bt} \; | \; \al \; \rgl  \lgl \; \bt \; |
\ee
is not a projector, since
$$
\hat P(z_m) \hat P(z_n) = 
\left(\; \sum_\al b^*_{m\al} b_{n\al} \;\right) \; | \; z_m \; \rgl \lgl \; z_n \; | \; .
$$

The prospect $A_n z_n$ is represented by the vector 
\be
\label{66}
| \; A_n z_n \; \rgl = | \; A_n \; \rgl \bigotimes |\; z_n \; \rgl = 
\sum_\al b_{n\al} |\; A_n \al \; \rgl
\ee
in the decision space $\mathcal{H}$. The prospect operator
\be
\label{67}
 \hat P(A_n z_n) \equiv |\; A_n z_n \; \rgl \lgl \; z_n A_n \; | = 
\hat P(A_n) \bigotimes \hat P(z_n) \;  ,
\ee
with the property
$$
 \hat P(A_m z_m)  \hat P(A_n z_n) = 
\dlt_{mn} \lgl \; z_n \; | \; z_n \; \rgl \hat P(A_n z_n) \; ,
$$
is not idempotent and is not a projector.

The resolution of unity
\be
\label{68}
\sum_n \hat P(A_n z_n) = 1
\ee
is understood in the weak sense as the equality on average
\be
\label{69}
\left\lgl \; \sum_n \hat P(A_n z_n) \; \right\rgl =  1 \;  ,
\ee
which implies
\be
\label{70}
 {\rm Tr}_\cH \; \hat\rho(t) \sum_n \hat P(A_n z_n) = 1 \; .
\ee
The extended version of the latter equality reads as
\be
\label{71}
 \sum_n \; \sum_{\al\bt} b^*_{n\al} b_{n\bt} \;
\lgl \; \al A_n \; | \; \hat\rho(t) \; | \; A_n\bt \; \rgl = 1 \; .
\ee
The set of the operators $\{\hat{P}(A_n z_n)\}$ forms a kind of the positive 
operator-valued measure \cite{Yukalov_53}.

Thus, making a decision on an alternative $A_n$, a decision maker, as a matter of fact,
considers a prospect $\pi_n = A_n z_n$, as far as, in addition to the rational
quantification of the alternative utility, this alternative is subject to irrational
subconscious evaluations. A decision maker not merely makes a decision on the usefulness 
of an alternative, but also experiences feelings of attraction or repulsion to this 
alternative. Thus the probability of choosing an alternative is in fact a behavioral 
probability of the associated prospect comprising a decision on the utility as well as 
on the attractiveness of the alternative \cite{Yukalov_49}. Keeping in mind that de facto 
one practically always considers prospects associated with the given alternatives, and in 
order not to complicate notation, we shall denote the behavioral probability of a prospect
\be
\label{72}        
p(\pi_n,t) \equiv {\rm Tr}_\cH \;\hat\rho(t) \; \hat P(\pi_n) \equiv p(A_n,t)
\ee
as identified with the alternative probability
\be
\label{73}
p(A_n,t) =  {\rm Tr}_\cH \; \hat\rho(t) \; \hat P(A_n z_n) \; .
\ee
The resolution of unity (\ref{70}) or (\ref{71}) guarantees the probability 
normalization
\be
\label{74}
\sum_n p(A_n,t) = 1 \; , \qquad 0 \leq p(A_n,t) \leq 1  \;  .
\ee

Separating in expression (\ref{73}) diagonal terms
\be
\label{75}
f(A_n,t) \equiv \sum_\al \; |\; b_{n\al} \;|^2 \;
\lgl \; \al A_n \; | \; \hat\rho(t) \; | \; A_n \al \; \rgl
\ee
from off-diagonal terms
\be
\label{76}
q(A_n,t) \equiv \sum_{\al\neq \bt} \;  b^*_{n\al} b_{n\bt} \;
\lgl \; \al A_n \; | \; \hat\rho(t) \; | \; A_n \bt \; \rgl \;  ,
\ee
we obtain the behavioral probability
\be
\label{77}
 p(A_n,t) = f(A_n,t) + q(A_n,t) \; .
\ee

In the theory of quantum measurements, the diagonal term corresponds to classical 
theory, while the second, off-diagonal, term is due to quantum interference and 
coherence. The transition from quantum to classical measurements and from quantum 
to classical probabilities is called decoherence \cite{Joos_54}. Thus the first term 
corresponds to a classical probability enjoying the properties
\be
\label{78}
 \sum_n f(A_n,t) = 1 \; , \qquad 0 \leq f(A_n,t) \leq 1  \;   .
\ee
In decision theory, the classical probability is responsible for a rational choice of
alternatives and can be named {\it rational fraction} or {\it utility fraction}, since 
its value is prescribed by the utility of the alternative 
\cite{Yukalov_18,Yukalov_30,Yukalov_49,Yukalov_55,Yukalov_56}.   

The second term, entirely due to quantum effects, has the properties
\be
\label{79}
\sum_n q(A_n,t) = 0 \; , \qquad -1 \leq q(A_n,t) \leq 1  \;
\ee
following from equations (\ref{77}) and (\ref{78}). The property (\ref{79}) is named the
alternation law \cite{Yukalov_18,Yukalov_30,Yukalov_49,Yukalov_55,Yukalov_56}. From 
normalization (\ref{74}), we also have
$$
 - f(A_n,t) \leq q(A_n,t) \leq 1 - f(A_n,t) \; .
$$
The quantum term $q$ in decision theory is associated with irrational feelings 
characterizing the emotional attitude of the decision maker to the quality of 
considered alternatives, because of which it can be called {\it irrational factor}, 
or {\it attraction factor}, or {\it quality factor} 
\cite{Yukalov_18,Yukalov_30,Yukalov_49,Yukalov_55,Yukalov_56}.   

As is seen, for an alternative $A_i$ to be stochastically preferred over $A_j$, so that 
$p(A_i) > p(A_j)$, it is not always sufficient to be more useful, but it is necessary to 
be sufficiently attractive. An optimal alternative is that possessing the largest 
probability among the set of the considered alternatives.  

For any generalization of decision theory, it is very important to include as a particular 
case the classical decision theory based on expected utility. Quantum decision theory 
\cite{Yukalov_18,Yukalov_30,Yukalov_49,Yukalov_55,Yukalov_56} satisfies this requirement. 
The return to classical decision theory happens when the quantum term $q$ tends to zero. 
This can be called the {\it quantum-classical correspondence principle}:
\be
\label{80}
 p(A_n,t) \ra f(A_n,t) \; , \qquad  q(A_n,t) \ra 0 \; .
\ee     
In the theory of quantum measurements, the disappearance of the quantum term, corresponding 
to quantum coherence, is called {\it decoherence}. The vanishing of this term can be due to 
the influence of surrounding environment \cite{Joos_54,Zeh_57,Joos_58,Zurek_59} or to the 
action of measurements \cite{Yukalov_61,Yukalov_60}. In decision theory, the decoherence can 
be caused by the influence of society providing information to the decision maker 
\cite{Yukalov_62}. In both these cases, random disturbances from either surrounding or 
measurement devices, lead to the irreversibility of time arrow \cite{Yukalov_63,Yukalov_64}.

\section{Evolution of Behavioral Probability}

The temporal evolution of the behavioral probability (\ref{77}) can be treated as in Sec. 3. 
The time dependence enters the matrix element
\be
\label{81}
 \lgl \; \al A_n \; | \; \hat\rho(t) \; | \; A_n \bt \; \rgl =
\sum_{n_1 n_2} U_{n_1\al}(t) \lgl \; A_n \; | \; n_1 \; \rgl 
\lgl \; \al n_1 \; | \; \hat\rho(0) \; | \; n_2 \bt \; \rgl \;
U^*_{n_2\bt}(t) \lgl \; n_2 \; | \; A_n \; \rgl \; .
\ee
If the process in the space of mind is asymptotically slow, then
$$
U_{n\al}(t) \simeq 1 \qquad ( g \ra 0 )
$$
and we get 
\be
\label{82}
 \lgl \; \al A_n \; | \; \hat\rho(t) \; | \; A_n \bt \; \rgl \simeq
 \lgl \; \al A_n \; | \; \hat\rho(0) \; | \; A_n \bt \; \rgl \; .
\ee
This means that
\be
\label{83}
  f(A_n,t) \simeq f(A_n,0) \; , \qquad  q(A_n,t) \simeq q(A_n,0) \; ,
\ee
so that the probability practically does not change:
\be
\label{84}
p(A_n,t) \simeq p(A_n,0) \qquad ( g \ra 0) \; .
\ee

In the opposite case of a fast process, when
$$
U_{m\al}(t) \; U^*_{n\bt}(t) \simeq \dlt_{mn} \dlt_{\al\bt} 
\qquad 
( g \ra \infty) \; ,
$$
we obtain
\be
\label{85}
 \lgl \; \al A_n \; | \; \hat\rho(t) \; | \; A_n \bt \; \rgl \simeq
\dlt_{\al\bt} \sum_m \; \lgl \; A_n \; | \; m \; \rgl
\lgl \; \al m \; | \; \hat\rho(0) \; | \; m \al \; \rgl \lgl \; m \; | \; A_n \; \rgl
\qquad ( g \ra \infty ) \;  .
\ee
Then the rational fraction reads as
\be
\label{86}
 f(A_n,t) \simeq \sum_\al \; | \; b_{n\al} \; |^2 \; 
\lgl \al A_n \; | \; \hat\rho(t) \; | \; A_n\al \; \rgl \;  ,
\ee
where
\be
\label{87} 
\hat\rho(t) = \sum_m \hat P_m \; \hat\rho(0) \; \hat P_m 
\qquad  
( g \ra \infty ) \; ,
\ee
while the quantum term, characterizing irrational feelings, vanishes:
\be
\label{88}
 q(A_n,t) \simeq 0 \qquad ( g \ra \infty ) \;   .
\ee
Therefore the behavioral probability reduces to the rational part,
\be
\label{89}
  p(A_n,t) \simeq  f(A_n,t) \qquad ( g \ra \infty ) \;  .
\ee
    
Successive decision making can be described similarly to Sec. 5. For two successive 
decisions, one has to consider the probability
\be
\label{90}
p(B_k A_n,t) = 
{\rm Tr}_\cH \; \hat\rho(t) \; \hat P(B_k z_k) \bigotimes \hat P(A_n z_n) \;  .
\ee

The above results demonstrate the essential dependence of the irrational term $q$ from the 
speed of taking decisions. The fact that, under a slow decision process, the probability 
practically does not vary, remaining close to that existing at the initial time, is not 
surprising. What is more interesting is that under a fast decision process the quantum term 
vanishes. This can be explained as follows. The quantum term is caused by the interference 
and entanglement of subconscious feelings in the subject consciousness. In order that these 
processes would happen they need some time. However in the case of an extremely fast decision 
making the subject just has no time for deliberation when the acts of coherence and 
entanglement could develop.

\section{Evaluation of Initial Probability}

Before considering any evolutional process, it is necessary to prescribe initial 
conditions corresponding to the initial moment of time that can be set as $t = 0$. The 
initial probability 
\be
\label{91}
p(A_n) \equiv p(A_n,0) = f(A_n) + q(A_n)
\ee
is formed by the initial rational fraction and irrational factor,
\be
\label{92}
f(A_n) \equiv f(A_n,0) \; , \qquad  q(A_n) \equiv q(A_n,0) \;  .
\ee

To estimate the value of the initial rational fraction, it is possible to use the Luce
rule \cite{Luce_65}, according to which, if the alternative $A_n$ is characterized by an 
attribute $a_n$, then the alternative weight can be written as
\be
\label{93}
 f(A_n) = \frac{a_n}{\sum_n a_n} \qquad ( a_n \geq 0 ) \; .
\ee
For concreteness, let us study the case where alternatives are represented by lotteries
\be
\label{94}
  A_n = \{ x_i , p_n(x_i) : \; i = 1,2, \ldots, N_n \}  \; ,
\ee
where $x_i$ are outcomes and $p_n(x_i)$ are the outcome probabilities normalized so that
$$
\sum_i p_n(x_i) = 1 \; , \qquad 0 \leq p_n(x_i) \leq 1 \; .
$$
Employing a utility function $u(x)$, one can define \cite{Neumann_9} the expected utility
\be
\label{95}
 U(A_n) = \sum_i u(x_i) p_n(x_i) \; .
\ee

The rational fraction describes the classical weight of an alternative that in the present 
case is connected with the expected utility in the following way \cite{Yukalov_55,Yukalov_56}. 
If the expected utilities of all alternatives are semi-positive, they can be associated with 
the alternative attributes,
\be
\label{96}
 a_n = U(A_n) \; , \qquad U(A_n) \geq 0 \; ,
\ee
while if all utilities are negative, the alternative attributes can be defined by the 
inverse quantities
\be
\label{97}
 a_n = \frac{1}{|\; U(A_n)\;|} \; , \qquad U(A_n) < 0 \;  .
\ee

In the case when among the expected utilities there are both positive as well as negative 
utilities, it is possible to use the shift taking account of the available wealth $U_0$. 
This is done in the following way. Finding the minimal negative utility
$$
 U_{min} \equiv \min_n \; U(A_n) < 0 \;  ,
$$
one defines 
$$
  U_0 \equiv | \; U_{min} \; | 
$$
interpreted as the wealth available to decision makers before they make decisions. 
Then the shifted utilities 
\be
\label{98}
  \overline U(A_n) \equiv U(A_n) + U_0 \geq 0 
\ee
are used instead of $U(A_n)$, and we return to the case of semi-positive expected 
utilities. 

The so defined rational fraction, or utility fraction, enjoys the natural properties: For
semi-positive utilities, the fraction tends to zero, when its utility tends to zero,
\be
\label{99}
 f(A_n) \ra 0 \; , \qquad   U(A_n) \ra + 0 \; ,
\ee
and tends to one, when its utility is very large,
\be
\label{100}
  f(A_n) \ra 1 \; , \qquad   U(A_n) \ra + \infty \;  .
\ee
For negative expected utilities, we have 
\be
\label{101}
  f(A_n) \ra 1 \; , \qquad   U(A_n) \ra - 0 \;  ,
\ee
and respectively,
\be
\label{102}
  f(A_n) \ra 0 \; , \qquad   U(A_n) \ra - \infty \;  .
\ee
When considering empirical data, the rational fraction characterizes the fraction of decision 
makers taking decisions on the basis of rational normative rules. A more elaborate expression 
for the rational fraction can be derived by employing the minimization of an information 
functional \cite{Yukalov_49,Yukalov_55,Yukalov_56}. 

The irrational factor $q$, as has been explained above, is a random quantity distributed 
over the interval $[-1,1]$. However, being a random variable does not preclude it to possess 
a typical average quantity. The ways of defining a non-informative prior for $q$ are described 
in Refs. \cite{Yukalov_30,Yukalov_49,Yukalov_55,Yukalov_56}. Below we present a slightly 
modified derivation of the non-informative prior for the irrational factor. 
 
Let the related distribution function be denoted as $\varphi(x)$. This distribution is 
normalized
\be
\label{103}
\int_{-1}^1 \vp(x)\; dx = 1 \;  .
\ee
Then the average value of a positive irrational factor is defined as
\be
\label{104}
 q_+ = \int_0^1 x \vp(x)\; dx \; .
\ee
Respectively, the average value of a negative factor is
\be
\label{105}
 q_- = \int_{-1}^0 x \vp(x)\; dx \;  .
\ee

\vskip 2mm

{\bf Proposition}. The non-informative priors for the average irrational factors are
\be
\label{106}
q_+ = \frac{1}{4} \; , \qquad q_- = -\; \frac{1}{4} \;  .
\ee
 
\vskip 2mm

{\it Proof}. With the notation
\be
\label{107}
\lbd_+ = \int_0^1 \vp(x)\; dx \;   , \qquad  \lbd_- = \int_{-1}^0 \vp(x)\; dx \; ,
\ee
the normalization condition (\ref{103}) takes the form
\be
\label{108}
\lbd_+ + \lbd_- =  1 \; , \qquad 0 \leq \lbd_\pm \leq 1 \; .
\ee
In the average irrational factors (\ref{104}) and (\ref{105}), $x$ is a monotonic function, 
while $\varphi(x)$ is integrable. Therefore employing the theorem of average yields
\be
\label{109}
q_+= x_+ \lbd_+ \; , \qquad  q_-= x_- \lbd_- \;  ,
\ee
where
\be
\label{110}
 0 \leq x_+ \leq 1 \; , \qquad - 1 \leq x_- \leq 0 \; .
\ee
As a non-informative prior for the values of $x_\pm$ and $\lambda_\pm$ one takes the averages 
over the domains of their definition, that is,
\be
\label{111}
 x_+ = \frac{1}{2} \; , \qquad x_- = -\; \frac{1}{2} \; , \qquad 
\lbd_\pm = \frac{1}{2} \;  .
\ee
Substituting this into quantities (\ref{104}) and (\ref{105}) proves equalities (\ref{106}). 
   
\vskip 2mm

This property is termed the {\it quarter law}. The mentioned above alternation law and the 
quarter law can be used for estimating the aggregate values of the behavioral probabilities 
according to the rule
\be
\label{112}
p(A_n) = f(A_n) \pm 0.25.
\ee
The sign of the irrational factor is prescribed in accordance to the alternative being 
attractive or repulsive \cite{Yukalov_30,Yukalov_49,Yukalov_55,Yukalov_56}.   

This approach was employed for explaining a number of paradoxes in classical decision 
making \cite{Yukalov_30,Yukalov_47,Yukalov_48,Yukalov_55,Yukalov_56,Yukalov_66,Yukalov_67} 
and was found to be in good agreement with a variety of experimental observations
\cite{Yukalov_49,Yukalov_55,Yukalov_56,Favre_68}.

\section{Quantum Intelligence Network}

In the previous sections, we have considered quantum decision making by subjects that act
independently of each other. Even then the probability of alternatives could vary with time 
due to the entangling properties of the evolution operator inducing entanglement in the 
decision space \cite{Yukalov_34,Yukalov_35,Yukalov_69}, in addition to that caused by the 
initial subject state of mind \cite{Yukalov_70}. The situation becomes much more involved 
when subjects exchange their information with each other. When there is a society of 
subjects interacting through the exchange of information, we obtain a network. Since the 
agents make decisions, we have an intelligence network. The probabilities of alternatives 
will vary owing to the informational interaction. From empirical investigations, it is 
known that decision makers do alter their decisions as a result of mutual charing of 
information \cite{Charness_71,Blinder_72,Cooper_73,Charness_74,Charness_75,Chen_76,
Charness_77,Kuhberger_78}. This type of intelligence networks plays an important role as 
prolegomena into the problem of artificial intelligence 
\cite{Pearl_79,Wooldridge_80,Russel_81}. 

Assume that we are considering a society of $N$ agents enumerated by the index 
$i=1,2,\ldots,N$, who decide with respect to alternatives $A_n$, with $n=1,2,\ldots,N_A$. 
So, the probability for an $i$-th agent deciding on an alternative $A_n$ at the moment of 
time $t$ is $p_i(A_n,t)$. Each probability satisfies the standard normalization
\be
\label{113}
 \sum_{n=1}^{N_A} \; p_i(A_n,t) = 1 \; , \qquad  0 \leq p_i(A_n,t) \leq 1 \; .
\ee
Accordingly, the related rational fraction $f_i(A_n,t)$ is  normalized as
\be
\label{114}
 \sum_{n=1}^{N_A} \; f_i(A_n,t) = 1 \; , \qquad  0 \leq f_i(A_n,t) \leq 1 \;  .
\ee
And for the irrational factor, one has
\be
\label{115}
 \sum_{n=1}^{N_A} \; q_i(A_n,t) = 0 \; , 
\qquad  
 -f_i(A_n,t) \leq q_i(A_n,t) \leq 1 -  f_i(A_n,t) \; .
\ee

To give a realistic description for the temporal evolution of an intelligence network, 
it is required to take into account that decision making needs some time after receiving 
information. Denoting this delay time as $\tau$, allows us to represent the dynamics of 
a behavioral probability as
\be
\label{116}
 p_i(A_n,t+\tau) = f_i(A_n,t) + q_i(A_n,t)  \; .
\ee
The rational fraction weakly varies with time, so that on the time scale shorter than the 
discounting time \cite{Loewenstein_82} it can be treated as constant. Its value is prescribed 
by the utilities of the given alternatives, as is explained in Sec. 8. 

The time dependence of the irrational factor can be derived following Sec. 7, similarly to 
the derivation of the coherent interference term in the theory of quantum measurements, where 
the role of irrational subconscious feelings is played by the random influence of environment 
\cite{Yukalov_60,Yukalov_61,Yukalov_63,Yukalov_64,Yukalov_83}. If a decision maker at time 
$t$ accumulates the amount of information $M_i(t)$ from other members of the society, then 
the irrational factor is discounted from its initial value, becoming equal to
\be
\label{117}
 q_i(A_n,t) = q_i(A_n,0) \exp\{ - M_i(t) \} \;  .
\ee
The amount of the accumulated information composes \cite{Yukalov_62,Yukalov_84} the 
information-memory functional
\be
\label{118}
M_i(t) = \sum_{t'=1}^t \; \sum_{j=1}^N \; J_{ij}(t,t') \mu_{ij}(t') \;  ,
\ee
in which $J_{ij}(t,t')$ is the interaction-memory function and $\mu_{ij}(t)$ is the 
information gain by a subject $i$ from a subject $j$ at time $t$. To exclude self-action,
one has to set either $J_{ii} = 0$ or $\mu_{ii} = 0$. At the initial moment of time there  
is no yet additional information, so that one has the initial condition
\be
\label{119}
 M_i(0 ) = 0  \; .
\ee
  
The information gain can be modeled by the Kullback–Leibler \cite{Kullback_85,Kullback_86}
relative information 
\be
\label{120}
 \mu_{ij}(t) = 
\sum_{n=1}^{N_A} \; p_i(A_n,t) \; \ln \; \frac{p_i(A_n,t)}{p_j(A_n,t)} \;  .
\ee
The information gain (\ref{120}) is semi-positive, $\mu_{ij}\geq 0$, due to the inequality 
$\ln x\geq 1-1/x$. As is evident, $\mu_{ii}=0$. Depending on the range of the interactions 
between the agents and the longevity of their memory, there can arise different situations 
whose typical examples are as follows.

{\it Long-range interactions} have the form
\be
\label{121}
  J_{ij}(t,t') = \frac{1}{N-1} \; J(t,t') \qquad ( i \neq j)  \; .
\ee

{\it Short-range interactions} act only between the nearest neighbors,
\be
\label{122}
 J_{ij}(t,t') = J(t,t') \dlt_{\lgl ij\rgl} \; ,
\ee
where $\delta_{<ij>}$ equals one for $i$ and $j$ being the nearest neighbours and is zero 
otherwise.

{\it Long term memory}, that lasts forever, does not depend on time,
\be
\label{123}
J_{ij}(t,t') = J_{ij} \;  .
\ee

{\it Short-term memory}, on the contrary, corresponds to the situation, when only the last 
step is remembered,
\be
\label{124}
 J_{ij}(t,t') = J_{ij} \dlt_{tt'} \; .
\ee

Combining these ultimate cases gives us the following four possibilities for the 
information-memory functional.

{\bf Long-range interactions and long-term memory}:
\be
\label{125}
 M_i(t) = \frac{J}{N-1} \sum_{t'=1}^t \; \sum_{j=1}^N \; \mu_{ij}(t') \; .
\ee

{\bf Long-range interactions and short-term memory}:
\be
\label{126}
 M_i(t) = \frac{J}{N-1}  \sum_{j=1}^N \; \mu_{ij}(t) \;  .
\ee

{\bf Short-range interactions and long-term memory}:
\be
\label{127}
  M_i(t) = J \sum_{t'=1}^t \; \sum_{j=1}^N \; \dlt_{\lgl ij\rgl} \mu_{ij}(t') \;  .
\ee

{\bf Short-range interactions and short-term memory}:
\be
\label{128}
 M_i(t) = J  \sum_{j=1}^N \; \dlt_{\lgl ij\rgl} \mu_{ij}(t) \;  .
\ee

Keeping in mind modern human societies, we have to accept that long-range interactions are 
more realistic. This is because the modern-day information exchange practically does not 
depend on the location of interacting agents who are able to exchange information through 
phone, Skype, Zoom, etc.

\section{Case of Two Alternatives}

A very frequent situation is when agents have to decide between two suggested alternatives, 
that is, when $N_A = 2$. Then it is straightforward to simplify the notation by setting
$$
p_i(A_1,t) \equiv p_i(t) \; , \qquad p_i(A_2,t) = 1 - p_i(t) \; ,
$$
$$
f_i(A_1,t) \equiv f_i(t) \; , \qquad f_i(A_2,t) = 1 - f_i(t) \; ,
$$
\be
\label{129}
 q_i(A_1,t) \equiv q_i(t) \; , \qquad q_i(A_2,t) =  - q_i(t) \;  .
\ee
Now the dynamics is governed by the equation
\be
\label{130}
 p_i(t + \tau) = f_i(t) + q_i(t) \; ,
\ee
with the irrational factor
\be
\label{131}
 q_i(t) = q_i(0) \exp\{ - M_i(t) \} \;  .
\ee
The information gain takes the form
\be
\label{132} 
 \mu_{ij}(t) = p_i(t) \; \ln \; \frac{p_i(t)}{p_j(t)} + 
[ \; 1 - p_i(t) \; ] \; \ln \; \frac{1-p_i(t)}{1-p_j(t)} \; .
\ee
As is seen, $\mu_{ii} = 0$. Setting initial conditions, it is necessary to obey the 
restriction
$$
- f_i(0) \leq q_i(0) \leq 1 - f_i(0) \;   .
$$
 
Another realistic simplification could be when the considered society consists of two types 
of agents essentially differing by their initial decisions. Then the question is: How the 
initial decisions would vary with time being caused by the mutual exchange of information? 
Will agents with different initial decisions come to a consensus, as it often happens after 
a number of interactions \cite{Centola_87,Centola_88}?   
 
When the society can be divided into two parts of typical agents, with each part having 
similar initial decisions within the group, but essentially different initial decisions 
between the groups, the situation becomes equivalent to the consideration of two typical 
agents with these different initial decisions. As is explained above, long-range interactions 
are more realistic for the information exchange in the modern society. Below we present the 
results of numerical calculations for this type of interactions. The behaviour of the society 
strongly depends on the type of memory the agents have.

\vskip 2mm

{\bf Long-term memory}. In the case of long-term memory, the information-memory functionals 
for two groups are
\be
\label{133}  
 M_1(t) = J \sum_{t'=1}^t \; \mu_{12}(t') \; , \qquad 
 M_2(t) = J \sum_{t'=1}^t \; \mu_{21}(t') \;  .
\ee

The rational fractions are kept constant in time and the parameters are set $J=1$ and 
$\tau=1$. The society dynamics is strongly influenced by the initial decisions. There can 
happen two types of behaviour depending on the relations between the initial rational 
fractions and irrational factors. 

(i) {\it Rational group conventions}. There is the rational-irrational accordance in the
initial choice of both groups. Then at the initial moment of time, one group, say the first 
group, estimates the utility of the first alternative higher than the second group. Taking 
account of irrational feelings keeps the same preference with respect to behavioral 
probabilities:
\be
\label{134}
 f_1(0) > f_2(0) \; ,  \qquad p_1(0) > p_2(0) \; .
\ee
Recall that $f_i(t) \equiv f_i(A_1,t)$ and $p_i(t) \equiv p_i(A_1,t)$. Respectively, if the 
first group estimates the utility of the first alternative lower than the second group, the 
irrational feelings do not change this preference:
\be
\label{135}
f_1(0) < f_2(0) \; , \qquad p_1(0) < p_2(0) \; .
\ee
In the case of this rational-irrational accordance, independently of initial conditions, the 
behavioral probabilities tend to the respective rational fractions:
\be
\label{136}
 p_i(t) \ra f_i(0) \qquad ( t \ra \infty ) \; .
\ee

(ii) {\it Common convention}. At the initial time, the inequalities between the rational 
fractions of the groups and the inequalities between their behavioral probabilities are 
opposite with each other, that is, one has either
\be
\label{137}
 f_1(0) > f_2(0) \; ,  \qquad p_1(0) < p_2(0) \;   ,
\ee
or 
\be
\label{138}
 f_1(0) < f_2(0) \; , \qquad p_1(0) > p_2(0) \;  .
\ee
In this case, the behavioral probabilities tend with time to the common convention:
\be
\label{139}
 p_i(t) \ra p^* \qquad ( t \ra \infty ) \;  
\ee
approximately equal to
\be
\label{140}
 p^* = \frac{f_1(0)q_2(0)-f_2(0)q_1(0)}{q_2(0)-q_1(0)} \;  .
\ee

These results can be interpreted in the following way. In the situation of the 
rational-irrational accordance, the decision makers are more rational, while irrational 
feelings, such as emotions, play less important role. Under the prevalence of rational
arguments, the agents are inclined to choose the alternative with a higher utility. 

In the case of the rational-irrational discordance, the decision makers are forced to more 
efficiently exchange information, as a result of which they manage to develop a mutual 
convention. 
   
\vskip 2mm

{\bf Short-term memory}. In this case, the information-memory functionals are
\be
\label{141}
M_1(t) = J \mu_{12}(t) \; , \qquad M_2 = J \mu_{21}(t) \;  .
\ee
We set again $J = 1$ and $\tau = 1$. Numerical solution of the evolution equations reveals
the existence of two types of possible dynamics.

(i) {\it Group conventions}. The behavioral probabilities for each group tend with time 
to their own limits not coinciding with the corresponding rational fractions:
\be
\label{142}
 p_i(t) \ra p^*_i \qquad ( t \ra \infty) \;  .
\ee

(ii) {\it Everlasting fluctuations}. The behavioral probabilities for both groups do not
tend to any fixed point, but demonstrate everlasting oscillations. The details of the above 
numerical solutions can be found in Ref. \cite{Yukalov_84}.     

In the society with short-term memory, there is no enough accumulated information for the
formation of a common convention. Each group in such a society either develops their own 
goal, not necessarily rational, or constantly fluctuates without elaborating a consensus.

\section{Dynamic Decision Inconsistencies}

There are several so-called dynamic or time inconsistencies in decision making 
characterizing situations in which a decision-maker's preferences change over 
time in such a way that a preference at one moment of time can become inconsistent 
with a preference at another point in time. Below we consider some of these 
inconsistencies and show that they find quite natural explanations in QDT.

\subsection{Question order bias}

The order that several questions are asked in a survey or study can influence the 
answers that are given as much as by $40 \%$ \cite{Lavrakas_105,Pew_106}. That is 
because the human brain has a tendency to organize information into patterns. The 
earlier questions may provide information that subjects use as context in formulating 
their subsequent answers, or affect their thoughts, feelings and attitude towards the 
questioned problem. Sociological research gives a number of illustrations of this 
bias \cite{Pew_106}. Thus from a December 2008 poll we know that when people were 
asked "All in all, are you satisfied or dissatisfied with the way things are going 
in this country today?" immediately after having been asked "Do you approve or 
disapprove of the way George W. Bush is handling his job as president?", $88$ percent 
said they were dissatisfied, compared with only $78$ percent without the context of 
the prior question. 

In classical probability theory, the probability of joint events is symmetric. 
This is contrary to the quantum decision theory. Assume an interrogator first asks 
a question $A$, so that the answer suggests two alternatives: yes ($A_1$) or no 
($A_2$). After this, the interrogator poses another question $B$, also with the 
possible answers: yes ($B_1$) or no ($B_2$). As follows from Sec. 5, the probability 
$p(A_i B_j)$ does not equal the probability $p(B_j A_i)$, since they are defined 
through different decision states. The situation is similar to that occurring in 
the problem of quantum contextuality \cite{Khrennikov_107,Khrennikov_108,Dzhafarov_109}, 
where two random variables ($A_i B_j$ and $B_j A_i)$ cannot be characterized by a 
single density matrix. 

In that way, taking account of dynamic evolution in QDT shows that, in general, the 
alternatives are not commutative, in the sense that $p(A_i B_j) \neq p(B_j A_i)$, 
which is in agreement with the empirical data.   

\subsection{Planning paradox}

In classical decision theory, there is the principle of dynamic consistency according 
to which a decision taken at one moment of time should be invariant in time, provided 
no new information has become available and all other conditions are not changed. Then 
a decision maker, preferring an alternative at time $t_1$ should retain the choice at 
a later time $t_2>t_1$. However this principle is often broken, which is called the 
effect of dynamic inconsistency. 

A stylized example of dynamic inconsistency is the planning paradox, when a subject 
makes a plan for the future, while behaving contrary to the plan as soon as time comes 
to accomplish the latter. A typical case of this inconsistency is the stop-smoking 
paradox \cite{Benfari_110,Westmaas_111}. A smoker, well understanding damage to health 
caused from smoking, plans to stop smoking in the near future, but time goes, future 
comes, however the person does not stop smoking. Numerous observations \cite{Benfari_110} 
show that $85\%$ of smokers do plan to stop smoking in the near future, however only 
$36\%$ really stop smoking during the next year after making the plan. It is possible 
to pose the question: would it be feasible to predict the percentage of subjects who 
will really stop smoking during the next year, knowing that at the present time $85\%$ 
of them plan doing this. Below we show that QDT allows us to make this prediction 
\cite{Yukalov_47}.  

Let us denote by $A_1$ the alternative to stop smoking in the near future, and by 
$A_2$, the alternative not to stop smoking. And let us denote by $B_1$ the decision 
to really stop smoking, while by $B_2$, the decision of refusing to really stop 
smoking. According to QDT, the corresponding probabilities are
$$
p(A_1) = f(A_1) + q(A_1) \; , \qquad p(A_2) = f(A_2) + q(A_2) \;  ,
$$
$$
p(B_1) = f(B_1) + q(B_1) \; , \qquad p(B_2) = f(B_2) + q(B_2) \; .
$$
The utility factors do not change in time, so that the utility of planning to stop 
in the near future is the same as the utility to stop in reality and the utility not 
to stop in the near future is the same as that of not stopping in reality,  
$$
f(A_1) = f(B_1) \; , \qquad f(A_2) = f(B_2) \;  .
$$
Planning to stop smoking, subjects understand the attractiveness of this due to health
benefits. Hence the average attraction factors, according to Sec. 8, are 
$$
 q(A_1) = \frac{1}{4} \; , \qquad q(A_2) = -\; \frac{1}{4} \;  .
$$
But as soon as one has to stop smoking in reality, one feels uneasy from the necessity 
to forgo the pleasure of smoking, because of which the attraction factors become
$$
q(B_1) = -\; \frac{1}{4} \; , \qquad q(B_2) = \frac{1}{4} \;   .
$$
This is to be complemented by the normalization conditions
$$
p(A_1) + p(A_2) = 1 \; , \qquad p(B_1) + p(B_2) = 1 \;  ,
$$
$$
f(A_1) + f(A_2) = 1 \; , \qquad f(B_1) + f(B_2) = 1 \;   .
$$
Since $85\%$ of subjects plan to stop smoking, we have
$$
p(A_1)  = 0.85 \; , \qquad  p(A_2) = 0.15 \;   .
$$ 
Solving the above equations, the fraction of subjects who will really stop smoking is 
predicted to be
$$
p(B_1)  = 0.35 \; , \qquad  p(B_2) = 0.65 \;   .
$$
This is in beautiful agreement with the observed fraction of smokers really stopping smoking 
during the next year after taking the decision \cite{Benfari_110},
$$ 
p_{exp}(B_1)  = 0.36 \; , \qquad  p_{exp}(B_2) = 0.64 \;   .
$$
Thus, knowing only the percentage of subjects planning to stop smoking, it is 
straightforward, by means of QDT, to predict the fraction of those who will stop 
smoking in reality. This case is also of interest because it gives an example of 
preference reversal: when planning to stop smoking, the relation between the 
probabilities is reversed as compared to the relation between the fractions of 
those who have really stopped smoking,
$$
 p(A_1) >p(A_2) \; , \qquad p(B_1) < p(B_2) \;  .
$$

\subsection{Disjunction effect}

Disjunction effect is the violation of the sure-thing principle \cite{Savage_112}. 
This principle states: {\it If the alternative $A_1$ is preferred to the alternative 
$A_2$, when an event $B_1$ occurs, and it is also preferred to $A_2$, when an event 
$B_2$ occurs, then $A_1$ should be preferred to $A_2$, when it is not known which of 
the events, either $B_1$ or $B_2$, has occurred}. This principle is easily illustrated 
for classical probability. Let $B=B_1+B_2$ be the alternative when it is not known 
which of the events, either $B_1$ or $B_2$, has occurred. For a classical probability, 
one has
$$
f(A_nB) = f(A_nB_1) + f(A_nB_2) \qquad ( n = 1,2 ) \;   .
$$
From here, it immediately follows that if  
$$
 f(A_1B_1) > f(A_2B_1) \; , \qquad   f(A_1B_2) > f(A_2B_2) \; ,
$$
then 
$$
f(A_1B) > f(A_2B) \; . 
$$

However, empirical studies have discovered numerous violations of the sure-thing 
principle, which was called disjunction effect \cite{Tversky_113}. Such violations 
are typical for two-step composite games of the following structure. First, a group 
of agents takes part in a game, where each agent can either win (event $B_1$) or lose 
(event $B_2$), with equal probability $0.5$. They are then invited to participate in 
a second game, having the right either to accept the second game (event $A_1$) or to 
refuse it (event $A_2$). The second stage is realized in different variants: One can 
either accept or decline the second game under the condition of knowing the result of 
the first game. Or one can either accept or decline the second game without knowing 
the result of the first game. The probabilities, as usual, are understood in the 
frequentist sense as the fractions of individuals taking the corresponding decisions 
\cite{Yukalov_84}. 

From the experiment of Tversky and Shafir \cite{Tversky_113} we have
$$
f(A_1 B_1) = 0.345 \; , \qquad f(A_1 B_2) = 0.295 \; ,  
$$
$$
f(A_2 B_1) = 0.155 \; , \qquad f(A_2 B_2) = 0.205 \; .
$$
This shows that the alternative $A_1 B$ is more useful than $A_2 B$, since
$$
f(A_1 B) = 0.64 \; , \qquad f(A_2 B) = 0.36 \; , 
$$
which seems to agree with the sure-thing principle. However $f(A_n B)$ is not yet the 
whole probability that reads as
$$
p(A_nB) = f(A_nB) + q(A_nB) \qquad ( n = 1,2) \;   .
$$

Taking a decision to play, without knowing the result of the first game, is less 
attractive, because of which $q(A_1 B) = -1/4$, while $q(A_2 B) = 1/4$. As a result, 
we find
$$
p(A_1B) = 0.39 \; , \qquad p(A_2B) = 0.61 \;   ,
$$
which is in good agreement with the empirical data of Tversky and Shafir 
\cite{Tversky_113},
$$
p_{exp}(A_1B) = 0.36 \; , \qquad p_{exp}(A_2B) = 0.64 \;   .
$$
 
The dynamic consistency of the disjunction effect can be analyzed as in Secs. 9 
and 10. Details can be found in Ref. \cite{Yukalov_84}, where it is shown that, 
when decision makers are allowed to exchange information, the absolute values of 
the attraction factors diminish with time. This conclusion also is in good agreement 
with empirical observations \cite{Kuhberger_114}.

\section{Dynamic Preference Intransitivity}

Transitivity is of central importance to both psychology and economics. It is 
the cornerstone of normative and descriptive decision theories \cite{Neumann_9}. 
Individuals, however, are not perfectly consistent in their choices. When faced
with repeated choices between alternatives $A$ and $B$, people often choose in 
some instances $A$ and $B$ in others. Such inconsistencies are observed even in 
the absence of systematic changes in the decision maker's taste, which might be 
due to learning or sequential effects. The observed inconsistencies of this type 
reflect inherent variability or momentary fluctuation in the evaluative process. 
Then preference should be defined in a probabilistic fashion \cite{Luce_65}. 
Nevertheless, there happen several choice situations where time transitivity may 
be violated even in a probabilistic form. Then one says that there is dynamic 
preference intransitivity \cite{Tversky_115,Fishburn_116,Fishburn_117,
Makowski_118,Makowski_119,Muller_120,Panda_121}.   

The occurrence of preference intransitivity depends on the considered decision 
model and on the accepted definition of transitivity. But the general meaning is as 
follows. Suppose, one evaluates three alternatives $A,\; B,\; C$, considering them 
in turn by pairs. One compares $A$ and $B$ and, according to the selected definition 
of preference, concludes that $A$ is preferred over $B$, which can be denoted as $A>B$. 
Then one compares $B$ and $C$, finding that $B>C$. Finally, comparing $C$ and $A$, 
one discovers that $C>A$. This results in the preference loop $A>B>C>A$ signifying 
the intransitivity effect. 

As a simple illustration of the intransitivity effect, we can adduce the Fishburn 
\cite{Fishburn_117} example. Imagine that a person is about to change jobs. When 
selecting a job, the person evaluates the suggested salary and the prestige of the 
position. There are three choices: job $A$, with the salary $65000\$ $, but low 
prestige; job $B$, with the salary $58000\$ $ and medium prestige; and job $C$, with 
the salary $50000\$ $ and high prestige. The person chooses $A$ over $B$ because of 
the better salary, while a small difference in prestige, $B$ over $C$ because of the 
same reason, and comparing $C$ and $A$, the person prefers $C$ because of the higher 
prestige, although a lower salary. Thus one comes to the preference loop $A>B>C>A$.

Let us show how this problem can be resolved in QDT. Recall that the definition 
of the behavioral probability (\ref{72}) is contextual, in the sense that it depends 
on the initial conditions for the decision state and on the given time. This means 
that the comparison of each pair of alternatives constitutes a separate contextual 
choice, even if the external conditions remain unchanged. The utility factors can 
be calculated as described in Sec. 8. Thus, considering the pair $A$ and $B$ in the 
Fishburn example, we have the utility factors
$$
f_1(A) = 0.528 \; , \qquad f_1(B) = 0.472 \;  ,
$$
where the label marks the moment of time $t_1$ and an initial condition 
$\hat{\rho}_1(t_1)$ for the decision state. Due to the close prestige of the both 
jobs, their attraction factors coincide, which, as follows from the alternation law 
in Sec. 6, gives 
$$
 q_1(A) = q_1(B) = 0 \;  .
$$
This leads to the behavioral probabilities
$$
p_1(A) = 0.528 \; , \qquad p_1(B) = 0.472 \;   .
$$
Since $p_1(A)>p_1(B)$, the job $A$ at time $t_1$ is stochastically preferred over $B$.

Similarly, comparing the jobs $B$ and $C$ at time $t_2$, we get the utility factors
$$
f_2(B) = 0.537 \; , \qquad f_2(C) = 0.463 \;  , 
$$
and the attraction factors
$$
 q_2(B) = q_2(C) = 0 \;   .
$$
Hence the probabilities are
$$
p_2(B) = 0.537 \; , \qquad p_2(C) = 0.463 \;   ,
$$
which implies that $B$ at time $t_2$ is stochastically preferred over $C$.

In the comparison of the jobs $C$ and $A$ at time $t_3$, we find the utility factors
$$
f_3(C) = 0.435 \; , \qquad f_3(A) = 0.565 \;   .
$$
Now the positions are of a very different quality, so that the attraction factors are
$$
q_3(C) = \frac{1}{4} \; , \qquad q_3(A) = -\; \frac{1}{4} \;   .
$$
Therefore the probabilities become
$$
p_3(C) = 0.685 \; , \qquad p_3(A) = 0.315 \;   .
$$
Then, at time $t_3$, the job $C$ is stochastically preferred over $A$.     
  
However, we should not forget that these comparisons were accomplished at different
moments of time and under different initial conditions. Therefore there is nothing 
extraordinary that differently defined probabilities can be intransitive. Even more,
there are arguments \cite{Tsetsos_122} that such intransitivities can be advantageous 
for alive beings in the presence of irreducible noise during neural information 
processing.      
       
The arising preference loops can be broken in two ways. First, the process of 
comparison requires time during which there can appear additional information. The 
attraction factors, as is explained in Secs. 9 and 10, vary with time, diminishing 
as time increases. When $q_3(C)$ and $q_3(A)$ tend to zero then 
$$
 p_3(C) ~ \longrightarrow ~ f_3(C) = 0.435 \;  ,
$$
$$
 p_3(A) ~ \longrightarrow ~ f_3(A) = 0.565 \; .
$$
Since now $p_3(C) < p_3(A)$, then $A$ becomes preferred over $C$, hence the preference
loop is broken. 

The other very natural way is as follows. As soon as there appears a preference loop, 
this implies that decisions at different moments of time and under different contexts 
should not be compared. One has to reconsider the whole problem at one given moment 
of time $t$. Considering all three alternatives $A, B, C$ in the frame of one given 
choice, we have
$$
f(A) = 0.376 \; , \qquad f(B) = 0.335 \; , \qquad f(C) = 0.289 \;   .
$$
The classification of the related qualities can be estimated according to the QDT 
rule
$$
q(A) = -\; \frac{1}{4} \; , \qquad q(B) = 0 \; , \qquad q(C) = \frac{1}{4} \;    .
$$
Then we find
$$
p(A) = 0.126 \; , \qquad p(B) = 0.335 \; , \qquad p(C) = 0.539 \;  ,
$$
which establishes the relation $(A \prec B \prec C)$ between all alternatives, and no 
problems or paradoxes arise.

\section{Conclusion}

The basic ideas of quantum decision theory are presented, with the emphasis on the problems 
of time evolution of decision processes. The relation between operationally testable events 
and behavioral features of taking decisions are elucidated. The interplay of rational and 
irrational sides of decision making is explained. The approach to describing quantum 
intelligence networks is developed. As illustrations, several time inconsistencies are
analyzed.

The main points of quantum decision theory are based on the techniques used in the theory 
of quantum measurements. Therefore the presented approach can be employed for 
characterizing evolutional processes in quantum measurements. The behaviour of many 
self-organizing complex systems is similar to decision making \cite{Yukalov_89}, because 
of which the approach could be useful in applications to complex systems and to the 
problems of artificial intelligence.

\section*{Acknowledgments}

The author is grateful to E.P. Yukalova for useful discussions.

\end{document}